Egger J., Zukic Dž., Bauer M. H. A., Kuhnt D., Carl B., Freisleben B., Kolb A., Nimsky Ch.

# A Comparison of Two Human Brain Tumor Segmentation Methods for MRI Data

**Abstract**

*The most common primary brain tumors are gliomas, evolving from the cerebral supportive cells. For clinical follow-up, the evaluation of the preoperative tumor volume is essential. Volumetric assessment of tumor volume with manual segmentation of its outlines is a time-consuming process that can be overcome with the help of computerized segmentation methods. In this contribution, two methods for World Health Organization (WHO) grade IV glioma segmentation in the human brain are compared using magnetic resonance imaging (MRI) patient data from the clinical routine. One method uses balloon inflation forces, and relies on detection of high intensity tumor boundaries that are coupled with the use of contrast agent gadolinium. The other method sets up a directed and weighted graph and performs a min-cut for optimal segmentation results. The ground truth of the tumor boundaries – for evaluating the methods on 27 cases – is manually extracted by neurosurgeons with several years of experience in the resection of gliomas. A comparison is performed using the Dice Similarity Coefficient (DSC), a measure for the spatial overlap of different segmentation results.*

**Introduction**

The most common primary brain tumors are gliomas, whereof 70% are among the group of malignant gliomas (anaplastic astrocytoma World Health Organization (WHO) grade III, glioblastoma multiforme (GBM) WHO grade IV) [11]. The GBM is one of the highest malignant human neoplasms. Due to the biological behavior, gliomas of WHO grade II to IV cannot be cured with surgery alone. The multimodal therapeutical concept involves maximum safe resection followed by radiation and chemotherapy, depending on the patient's Karnofsky[1] scale. The survival rate is still only approximately 15 months [12], despite new technical and medical accomplishments such as multimodal navigation during microsurgery, stereotactic radiation or the implementation of alkylating substances. Although there is still a lack of Class I evidence, literature today favors a maximum extent of resection in low- and high-grade gliomas as a positive predictor for longer patient survival [13]. The clinical follow-up of tumor volume is essential for an adaptation of the therapeutical concept. Therefore, the exact evaluation is fundamental to reveal a recurrent tumor or tumor progress as early as possible. Volumetric assessment of a tumor with manual segmentation of its outlines is a time-consuming process that can be overcome with the help of segmentation methods.

Others working on segmentation methods for brain tumors are Szwarc et al. [17]. They present a segmentation method of brain tumors in magnetic resonance imaging (MRI) data using fuzzy clustering techniques. Tests of their method have been performed on six MRI datasets of three subjects with high-grade astrocytomas. A comparison with manual segmentations verified by a radiologist resulted in Dice Similarity Coefficient (DSC) values from 67.21% to 75.63%. For a recent comprehensive overview of brain tumor segmentation methods based on MRI data, see the work of Angelini et al. [1]. Gibbs et al. [8] presented a method that combines thresholding, morphological edge detection and region growing. The method is manually initialized by providing an average gray value of the tumor and non-tumor surroundings. It is applied on enhanced T1-weighted MR images, slice by slice (consecutive slices initialized by final result of previous slice). Although they had slower hardware, their method took about ten minutes to execute. Letteboer et al. [14] proposed an interactive segmentation method and evaluated it on 20 clinical datasets. They rely on manual tracing of an initial slice, and apply morphological filter operations to divide the data in homogenous regions. However, their improvement over a completely manual segmentation with respect to segmentation speed is only threefold. A deformable model depending on intensity-based pixel probabilities for tumoral tissue has been introduced by Droske et al. [5]. They used a level set formulation, in order to split the MRI data into regions of similar image properties for tumor segmentation. The method was performed on image data of twelve patients. Since the method is interactive, the segmentation time is about three minutes. Clark et al. [3] proposed a knowledge-based automated segmentation on multispectral (T1, T2, PD) data to partition glioblastomas. After a training phase with fuzzy C-means classification, clustering analysis and a brain mask computation, initial tumor segmentation from vectorial histogram thresholding is post-processed to eliminate non-tumor voxels. The introduced system has been trained on three MRI volume datasets and has been tested on 13 unseen volume datasets. Segmentation based on outlier detection in T2-weighted images, with the ability to augment it with T1 image, has been developed by Prastawa et al. [15]. The image data is registered with a brain atlas, in order to employ distribution probabilities for different tissue classes. Voxels are statistically classified, and then snakes are applied to refine the segmentation. A constraint that edema is close to tumor is used to reduce false positives. The method is automatic after the user sets some general parameters. However, they have applied the method only to three real MRI datasets. For each case, the required time for automatic segmentation was about 90 minutes.

---

[1] Scale of 0-100, indicating functional impairments of the patient (0=dead, 100=healthy)

Sieg et al. [16] have presented a method for segmenting tumors using registered MRI images of types T1, T1 plus contrast enhancement, T2 and PD. They trained multilayer feed-forward neural network for voxel-oriented classification, and then chose the largest connected component as the tumor. They tested it on 22 images, but no computational time was provided.

In this contribution, two methods for WHO grade IV glioma segmentation in the human brain are compared using MRI patient data from the clinical routine. The first method uses balloon inflation forces, and relies on detection of high intensity tumor boundaries that are coupled with the use of contrast agent gadolinium. The second method sets up a directed and weighted graph and performs a min-cut for optimal segmentation results. The comparison is performed using the DSC, a measure for the spatial overlap of different segmentation results.

The paper is organized as follows: Section 2 presents the details of the proposed approaches. In Section 3, experimental results are presented. Section 4 discusses the paper and outlines areas for future work.

**Material and Methods**

**Preprocessing**

For the segmentation process of the pathologies (GBM WHO grade IV), we used 1.5 Tesla magnetic resonance imaging scans from the clinical routine. For the glioblastomas, we chose T1-weighted images after gadolinium-enhancement (mostly axial). The segmentation-outlines were marked by the contrast-enhanced structures. Both methods need a specific user initialization before the automatic segmentation can be performed. For the method that is based on balloon inflation forces the user draws an approximate outline on a slice that is nearly in the middle of the tumor (Figure 1 left). The graph-based method needs a user-defined seed point that is located approximately in the center of the tumor (Figure 1 right).

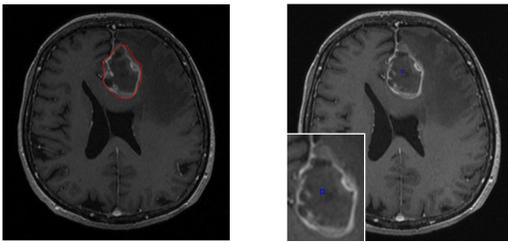

Fig. 1: Initialization for balloon inflation forces approach (left) and graph based approach (right)

**Segmentation with balloon inflation forces**

The main idea is to start with a small triangular surface mesh in the shape of a convex polyhedron at the approximate center of the glioma. Balloon inflation forces [4] are then used to expand this mesh iteratively, in which each iteration step consists of:

- Splitting of long edges (mesh refinement)
- Computation of surface normals per vertex and estimation of curvature
- Inflation (moving vertices outwards)
- Slight smoothing of the mesh

Vertices are moved outwards only if they are going to be placed into an area of similar or higher intensity (thus stopping once bright gadolinium-enhanced boundary has been reached, followed by lower intensities). Vertices with lower curvature are moved outwards by a larger amount, thus stimulating smoother meshes. And finally, vertices with high angle between normal and center-vertex-vector are inflated by a smaller amount, in order to penalize protrusions (Figure 2).

Smoothing the mesh is required to overcome the noise, which would otherwise cause many vertices to get stuck far away from the boundary. The segmentation is declared as finished when the inflation speed becomes slow (due to most vertices becoming fixed in the boundary).

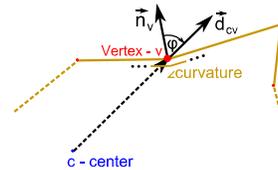

Fig. 2: Features involved in vertex movement calculation ($\vec{n}_v$ - normal at vertex v, $\vec{d}_{cv}$ - center-vertex-vector)

**Graph-based segmentation**

The overall graph-based method starts by setting up a directed 3D graph from a user-defined seed point that is located inside the object. To set up the graph, the method samples along rays that are sent through the surface points of a polyhedron with the seed point as center (Figure 3). The sampled points are the nodes $n \in V$ of the graph $G(V,E)$ and $E$ is a corresponding set of edges $e \in V$. There are edges between the nodes and edges that connect the nodes to a source node $s$ or a sink node $t$ to allow the computation of an $s$-$t$ cut (source and sink are virtual nodes). For details about setting up the graph, see [6,7].

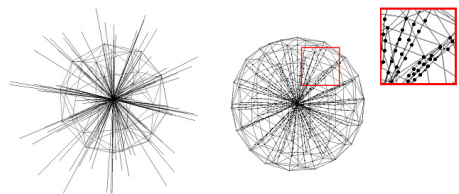

Fig. 3: Principle of sending rays through the surface of a polyhedron with 32 vertices (left). Sampling of the nodes for the graph along the rays (right)

After graph construction, the minimal cost closed set on the graph is computed via a polynomial time $s$-$t$ cut [2]. The $s$-$t$ cut creates an optimal segmentation of the object under the influence of the parameter $\Delta_r$ that controls the

Egger J., Zukic Dž., Bauer M. H. A., Kuhnt D., Carl B., Freisleben B., Kolb A., Nimsky Ch.

stiffness of the surface. A delta value $\Delta_r$ of 0 ensures that the segmentation result is a sphere. The weights $w(x,y,z)$ for every edge between $v \in V$ and the sink or the source node are assigned in the following manner: weights are set to $c(x,y,z)$ when $z$ is zero and otherwise to $c(x,y,z)-c(x,y,z-1)$, where $c(x,y,z)$ is the absolute value of the intensity difference between an average gray value of the desired object and the gray value of the voxel at position $(x,y,z)$.

**Postprocessing**

After having segmented the brain tumors, both methods call for the same post-processing steps. The resulting contours (given as point clouds) of the tumor boundaries have to be triangulated to get a closed surface. This closed surface is used to generate a solid 3D mask (representing the segmented tumor), which is done by voxelization of the triangulated mesh. With the resulting masks, three evaluation measures can be obtained: The tumor volume (cm$^3$), the number of voxels and the DSC [18]. The DSC is the relative volume overlap between A and R, where A and R are the binary masks from the automatic (A) and the reference (R) segmentation. V(•) is the volume (in cm$^3$) of voxels inside the binary mask, by means of counting the number of voxels, then multiplying with the voxel size.

**Results**

The presented method using balloon inflation forces has been implemented in C++. The segmentation took about 1 second per dataset on an Intel Core i7-920 CPU, 2.66 GHz (4 cores), on Windows7 x64. The graph-based approach has been implemented in C++ within the MeVisLab platform (http://www.mevislab.de/). Using 2432 and 7292 polyhedra surface points, the overall segmentation (sending rays, graph construction and min-cut computation) took less than 5 seconds on an Intel Core i5-750 CPU, 4x2.66 GHz, 8 GB RAM, Windows XP Professional x64 Version, Version 2003, SP 2. Manual segmentation took 6.93±4.11 minutes (minimum 3 minutes and maximum 19 minutes). To evaluate the approaches, neurological surgeons with several years of experience in the resection of tumors performed manual slice-by-slice segmentation of 27 selected WHO grade IV gliomas in T1-weighted contrast enhanced MRI datasets. The tumor outlines for the segmentation were displayed by the contrast enhancing areas. Afterwards, the segmentation results were compared with both presented methods via the presented evaluation measure (DSC). Table I shows the volume of tumor in cm$^3$, and the number of voxels of the manual segmentation and the two approaches. Additionally, the DSC for both approaches is denoted in Table I.

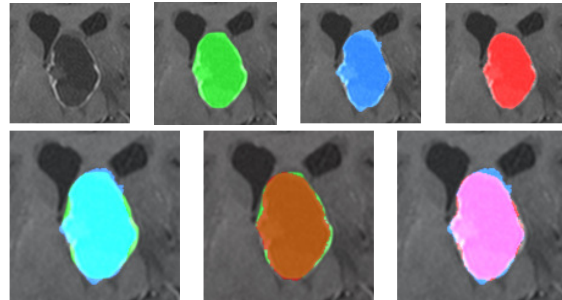

Fig. 4: First row: Close-up of one slice of the original image (left). Segmented tumor: manual (green), balloon inflation forces (blue) and graph-based (red). Second row – overlaps: cyan = manual + inflation forces, brown = manual + graph-based, pink = inflation forces + graph-based

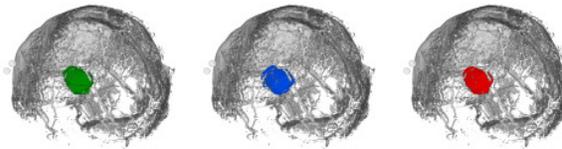

Fig. 5: Segmented tumor: manual segmentation (left), segmentation with balloon inflation forces (middle) and graph-based segmentation (right)

For a direct comparison of the presented approaches, Figure 4 shows axial MRI slices of a tumor, enhanced with the different segmentation results. In the upper row, from left to right: "native" slice without any segmentation, the manual segmentation, the balloon inflation forces based method, and the graph-based method, with the MRI slice enhanced with different segmentation masks, are shown. In the lower row, the manual mask has been superimposed with the balloon approach (left) and the graph-based approach (middle). Furthermore, the resulting segmentation masks from both approaches are blended in (right). The manual segmentation for this dataset took 4 minutes (139670 voxel and 16.26 cm$^3$). The automatic segmentation with balloon inflation forces (142469 voxel and 16.59 cm$^3$) yielded to a DSC of 89.50% and the graph-based approach (133254 voxel and 15.51 cm$^3$) provided a DSC of 93.43%. Figure 5 shows a manually segmented tumor as a 3D mask (left), the same tumor segmented with the balloon inflation forces based method (middle) and with the graph-based method (right).

TABLE I
SUMMARY OF RESULTS: MIN., MAX., MEAN AND STANDARD DEVIATION FOR BOTH APPROACHES

|  | Volume of tumor (cm$^3$) | | | Number of voxels | | | DSC$_{balloon}$ (%) | DSC$_{graph}$ (%) |
|---|---|---|---|---|---|---|---|---|
|  | manual | balloon | graph | manual | balloon | graph |  |  |
| min | 0.79 | 0.77 | 0.99 | 1526 | 2465 | 993 | 63.72 | 69.82 |
| max | 73.45 | 73.32 | 59.18 | 550307 | 446560 | 591801 | 94.02 | 93.82 |
| $\mu \pm \sigma$ | 21.64 ± 19.16 | 20.25 ± 19.27 | 20.64 ± 18.93 | 100349.9 | 86589.7 | 100222.9 | 80.46 ± 7.42 | 82.49 ± 8.19 |

## Discussion

In this contribution, two approaches for WHO grade IV glioma segmentation have been presented, evaluated and compared against each other. One method uses balloon inflation forces and relies on the detection of high-intensity tumor boundaries that are coupled with the use of contrast agent gadolinium. The other method sets up a directed and weighted graph and performs a min-cut for optimal segmentation results. The presented approaches have been compared and evaluated on various MRI datasets with WHO grade IV gliomas. Experts (neurosurgeons) with several years of experience in the resection of gliomas extracted the tumor boundaries manually to obtain the ground truth for the given data. The manually segmented results and the segmentation results of the presented approaches have been compared by calculating the average DSC.

In the case of gaps in the tumor border – due to lacking contrast agent, for example – the balloon inflation approach can provide a smoother (approximate) border in these missing parts. This is due to the iterative segmentation process that slowly fits step by step to the tumor border. However, iterative segmentation methods can always get stuck in local minima during the iterative (expansion) process. In contrast, a graph-cut approach always provides an optimal segmentation for the constructed graph. To get a precise tumor volume it is essential to develop methods – like introduced in this paper – which use all slices to calculate the tumor boundaries. Simpler methods like geometric models provide only a rough approximation of tumor volume and should not be used, as accurate determination of size is of paramount importance in order to draw safe conclusions in oncology [10].

There are several areas of future work. For example, the presented segmentation schemes should be enhanced with statistical information about shape and texture of the desired object [9]. Furthermore, the methods should be evaluated on MRI datasets with WHO grade I, II and III gliomas. Additionally, we work on an extensive evaluation of the user initialization and its impact on the automatic segmentation result: manual drawn outline for the balloon inflation approach and the user-defined seed-point for the graph-based approach.

## Acknowledgements

The authors would like to thank Fraunhofer MeVis in Bremen, Germany, for their collaboration and especially Horst K. Hahn for his support.

## Affiliation of the first Author

Dr. Jan Egger
University of Marburg, Department of Neurosurgery
Baldingerstraße, 35033 Marburg, Germany
Phone +49 6421 58 66754
egger@med.uni-marburg.de